\documentclass[conference]{IEEEtran}
\IEEEoverridecommandlockouts
\usepackage{cite}
\usepackage{amsmath,amssymb,amsfonts}
\usepackage{algorithmic}
\usepackage{graphicx}
\usepackage{multirow}
\usepackage{textcomp}
\usepackage{xcolor}
\def\BibTeX{{\rm B\kern-.05em{\sc i\kern-.025em b}\kern-.08em
    T\kern-.1667em\lower.7ex\hbox{E}\kern-.125emX}}
\begin{document}

\title{Integrating Distribution Matching into Semi-Supervised Contrastive Learning for Labeled and Unlabeled Data\\


}

\author{\IEEEauthorblockN{1\textsuperscript{st} Shogo Nakayama, Masahiro Okuda}
\IEEEauthorblockA{\textit{Doshisha University} \\
Kyoto, Japan\\
nakayama@vig.doshisha.ac.jp}}

\maketitle

\begin{abstract}
The advancement of deep learning has greatly improved supervised image classification. However, labeling data is costly, prompting research into unsupervised learning methods such as contrastive learning. In real-world scenarios, fully unlabeled datasets are rare, making semi-supervised learning (SSL) highly relevant in scenarios where a small amount of labeled data coexists with a large volume of unlabeled data. A well-known semi-supervised contrastive learning approach involves assigning pseudo-labels to unlabeled data. This study aims to enhance pseudo-label-based SSL by incorporating distribution matching between labeled and unlabeled feature embeddings to improve image classification accuracy across multiple datasets.
\end{abstract}

\begin{IEEEkeywords}
Semi-Supervised Learning, Contrastive Learning, Pseudo Labeling, Distribution Matching, Deep Learning, Image Classification
\end{IEEEkeywords}

\section{Introduction}
The development of semi-supervised learning (SSL) has the potential to reduce the burden of data labeling, which is a critical challenge in deep learning applications. SSL is particularly beneficial in tasks with limited labeled data, such as medical image analysis. One prominent SSL method, proposed by Lee et al. \cite{pseudo}, assigns pseudo-labels to unlabeled data by selecting the class with the highest prediction confidence. This approach allows models to make better use of unlabeled data but is prone to confirmation bias, as incorrect pseudo-labels can reinforce erroneous predictions. To mitigate this issue, an advanced method, FixMatch \cite{fixmatch} refines the pseudo-labeling approach by introducing a confidence threshold, ensuring that only high-confidence samples contribute to training. However, a significant limitation of FixMatch is that it discards samples below the confidence threshold, leading to inefficient utilization of valuable unlabeled data. To address this inefficiency, researchers have explored various enhancements to SSL frameworks. Hierarchical self-regularization \cite{hie} has been proposed as a means to optimize SSL performance by aligning feature distributions of labeled and unlabeled data. Furthermore, Gauffre et al. \cite{fix_con} integrated supervised contrastive learning with FixMatch, allowing the model to leverage low-confidence samples effectively. This integration helps mitigate the drawback of FixMatch by ensuring that all unlabeled samples contribute to learning, albeit with varying degrees of influence. Building upon these advances, this study proposes a novel approach that combines the semi-supervised contrastive learning framework \cite{fix_con} with the distribution matching technique used in hierarchical self-regularization. By aligning the feature distributions of labeled and unlabeled data, our method aims to fully exploit the information contained in unlabeled samples, thereby enhancing classification performance. The proposed framework not only improves the efficiency of pseudo-labeling but also ensures that even low-confidence samples contribute meaningfully to training. While semi-supervised contrastive learning allows the use of unlabeled data, low-confidence samples are assigned lower weights, limiting their impact. By integrating distribution matching, our method effectively aligns feature distributions between labeled and unlabeled data, enabling a more robust use of unlabeled samples and further enhancing classification performance.

The paper is structured as follows. Section 2 reviews related work on contrastive learning and SSL. Section 3 presents the proposed approach, detailing how semi-supervised contrastive learning is integrated with distribution matching. Section 4 describes the experimental setup including datasets and baseline models, followed by an evaluation of classification performance. Finally, Section 5 presents our conclusions.

\section{Related Work}

\subsection{Contrastive learning}

Contrastive learning has gained attention as an effective unsupervised learning technique. It learns feature representations by maximizing similarity between positive pairs and minimizing similarity between negative pairs. A notable example is SimCLR \cite{SimCLR}, which uses data augmentation to generate positive pairs from the same image. SimSiam \cite{Siam} eliminates negative pairs while maintaining performance. MoCo \cite{MoCo} dynamically manages negative pairs with a queue and utilizes momentum updates for stability. These contrastive learning methods have also been extended to supervised settings \cite{SupCon}, where pairs are formed based on class labels.

\subsection{Semi-Supervised Learning}
Semi-supervised learning involves the use of both labeled and unlabeled data. FixMatch \cite{fixmatch} assigns pseudo-labels based on confidence thresholds, but discards low-confidence samples. FlexMatch \cite{Flex} improves FixMatch by adaptively adjusting the confidence thresholds. MixMatch \cite{Mix} assigns smoothed pseudo-labels by averaging multiple predictions. While effective, these methods either discard some unlabeled data or increase computational costs. Hierarchical self-regularization \cite{hie} utilizes Maximum Mean Discrepancy (MMD) to align feature distributions of labeled and unlabeled data, enhancing SSL performance.

In \cite{fix_con}, although unlabeled data is incorporated into the training process, it is assigned a lower weight and thus not fully utilized. In \cite{hie}, MMD is used for the alignment of the distribution between the labeled and unlabeled data. However, the loss function is based on cross-entropy, which typically only considers labeled data or unlabeled data with high-confidence pseudo-labels. As a result, low-confidence unlabeled samples are not involved in the learning process. By integrating the strengths of both approaches, this study enables more effective utilization of unlabeled data through the combination of semi-supervised contrastive learning and distribution matching via MMD.

\section{Proposed Method}
\begin{figure*}[h]
\centering
\centerline{\includegraphics[width=0.7\linewidth]{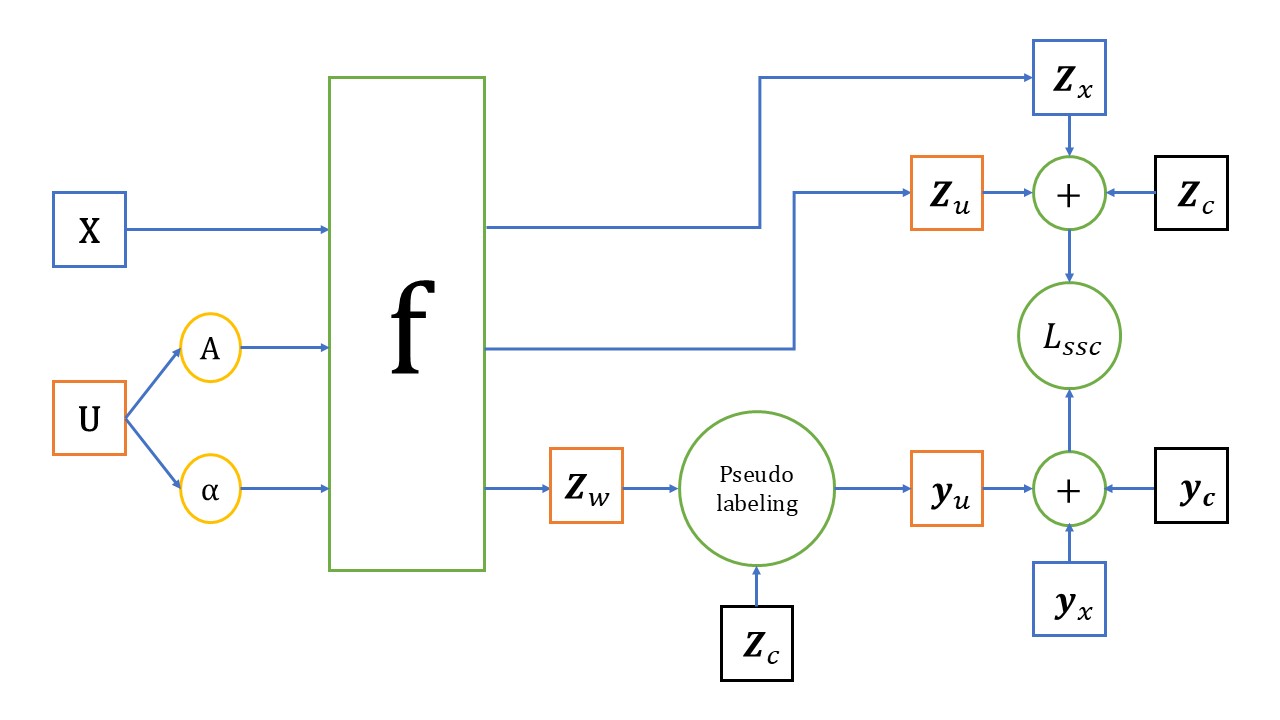}}
\caption{Overview of semi-supervised contrastive learning}
\label{fig_1}
\end{figure*}

\begin{figure}[htbp]
\centerline{\includegraphics[width=1\linewidth]{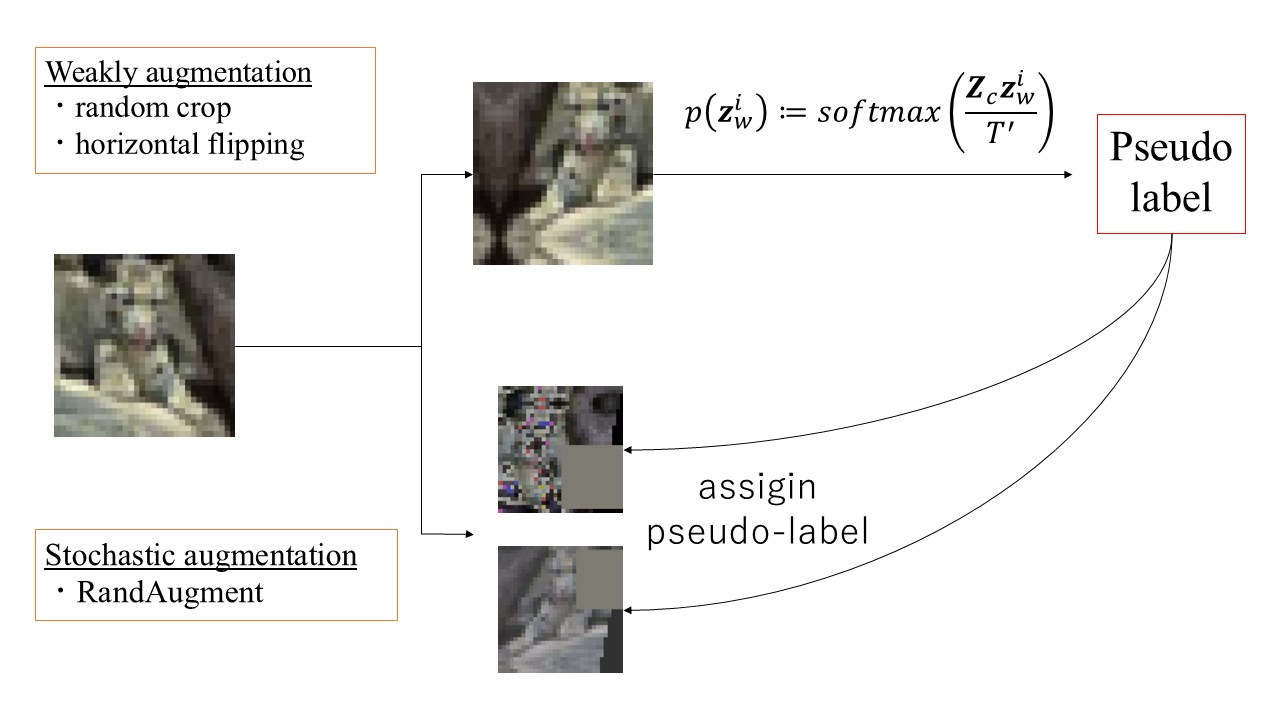}}
\caption{Pseudo Labeling ($p(\mathbf{z}_w^i)>\tau$)}
\label{pseudo}
\end{figure}

\begin{figure}[htbp]
\centerline{\includegraphics[width=1\linewidth]{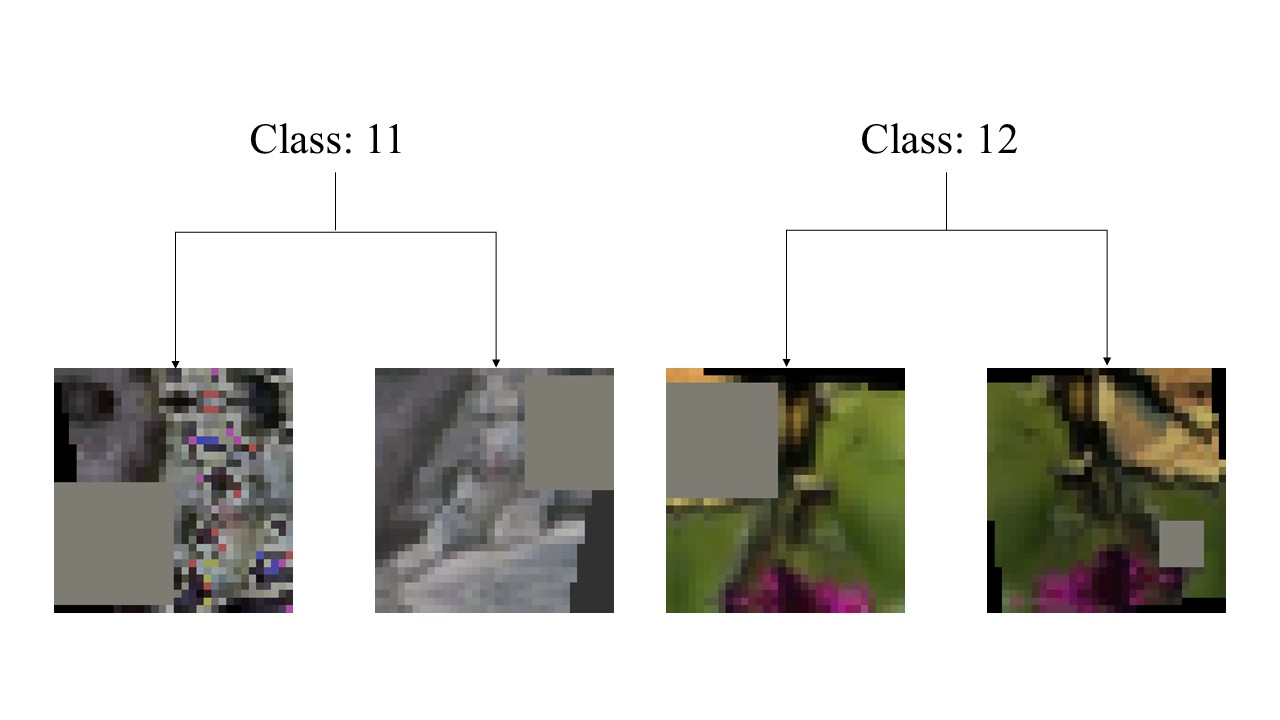}}
\caption{Pseudo Labeling ($p(\mathbf{z}_w^i)<\tau$), on CIFAR-10}
\label{pseudo_tau}
\end{figure}

In this chapter, we outline the method we used in our experiments. Our study adopts the semi-supervised contrastive learning approach introduced in \cite{fix_con}. The overall framework of \cite{fix_con} is illustrated in Figure \ref{fig_1}. Furthermore, we integrate the distribution matching technique for feature embeddings of labeled and unlabeled data.

\subsection{Semi-Supervised Contrastive Learning}\label{semi}

The conventional FixMatch method had challenges in effectively utilizing unlabeled data. By employing the semi-supervised contrastive learning method, it becomes possible to make full use of all the data.
\begin{equation}
\scalebox{1}{$
\mathbf{X} =
\begin{bmatrix}
\mathbf{x}^1 
\cdots 
\mathbf{x}^B
\end{bmatrix}
,
\mathbf{y}_x =
\begin{bmatrix}
y_x^1 
\cdots 
y_x^B
\end{bmatrix}
,
\mathbf{U} =
\begin{bmatrix}
\mathbf{u}^1 
\cdots 
\mathbf{u}^{\mu B}
\end{bmatrix}
$}
\end{equation}

Here, $\mathbf{X}$ represents a mini-batch of labeled data, and B denotes the mini-batch size of the labeled data. $\mathbf{y}_x$ corresponds to the class labels for each labeled data sample. $\mathbf{U}$ represents a mini-batch of unlabeled data, and $\mu$ denotes the ratio of the mini-batch size of unlabeled data to that of labeled data.

\scalebox{1}{\\
$\mathbf{Z}_x = f(\mathbf{X}) = \begin{bmatrix}
    \mathbf{z}_x^1,
    \cdots,
    \mathbf{z}_x^B
\end{bmatrix}^\top$}: represents the feature embeddings of labeled data. $f$ is an encoder that maps image data into a feature space.

\scalebox{1}{\\
$\mathbf{Z}_u = \begin{bmatrix}
    \mathbf{Z}_{s1} \\
    \mathbf{Z}_{s2}
\end{bmatrix} = \begin{bmatrix}
    f(A(\mathbf{U})) \\
    f(A(\mathbf{U}))\end{bmatrix}
    $}: represents two feature embeddings generated by applying probabilistic strong image augmentation to unlabeled data. $A$ is the application of strong image augmentations. For strong augmentations, RandAugment \cite{rand} is employed.

\scalebox{1}{\\
$\mathbf{Z}_w = f(\alpha(\mathbf{U}))= 
\begin{bmatrix}
\mathbf{z}_w^1,
\cdots,
\mathbf{z}_w^{\mu B}
\end{bmatrix}$}: represents the feature embeddings generated by applying weak image augmentation to unlabeled data. $\alpha$ is the application of weak image augmentations. Weak augmentations include random horizontal flipping and random cropping.

\scalebox{1}{\\
$\mathbf{Z}_c = \begin{bmatrix}
    \mathbf{z}_c^1,
    \cdots,
    \mathbf{z}_c^k
\end{bmatrix}$}
: represents prototypes for each class $k$.

\scalebox{1}{\\
$\mathbf{y}_c = \begin{bmatrix}
    y_c^1,
    \cdots,
    y_c^k
\end{bmatrix}$}
: represents the class labels for prototypes.

We describe the process of assigning pseudo-labels to unlabeled data. As shown in the following equation, the classification probability for each class is determined based on cosine similarity.

\begin{equation}
\scalebox{1}{$
p(\mathbf{z}_w^i) := \text{softmax}(\frac{\mathbf{Z}_c \mathbf{z}_w^i} {T'})
$}
\end{equation}

Here, $T'$ denotes the temperature used during pseudo-labeling.
The cosine similarity between the prototype of each class and the weakly augmented image is computed to obtain the classification probability for each class. If the highest classification probability exceeds a threshold $\tau$, the corresponding class is assigned as the pseudo-label. For unlabeled data that don't exceed the threshold, a unique label is assigned to each instance. The pseudo-label for each unlabeled sample is defined as follows.

\begin{equation}
\begin{aligned}
q = arg\max p(\mathbf{z}_w^i)\quad\quad\quad\quad
\\
y_{u}^i =
\begin{cases} 
y_{u}^{\uparrow i}=\begin{bmatrix}
    q,q
\end{bmatrix} & \text{if } \max p(\mathbf{z}_{w}^i) > \tau \\
y_{u}^{\downarrow i} + K & \text{otherwise}
\end{cases}
\end{aligned}
\end{equation}

Here, $\mathbf{K}$ represents the total number of classes (for example, in the CIFAR-10 dataset, $\mathbf{K}$ = 10). The overall pseudo-labeling procedure, including both high- and low-confidence cases, is illustrated in Figures \ref{pseudo} and \ref{pseudo_tau}.

\begin{table*}[t]
    \centering
    \caption{
    Comparison of accuracy. The highest accuracy for each condition is highlighted in bold. \\
    (1) \textnormal{\MakeLowercase{base}} represents the baseline model. \\
    (2) \textnormal{\MakeLowercase{w.mmd}} represents the model combining the baseline with MMD-based distribution matching.
}
    \label{tab:accuracy_comparison}
    \begin{tabular}{|l|cc|cc|cc|}
        \hline
        \textbf{Dataset} & \multicolumn{2}{c|}{\textbf{CIFAR-10}} & \multicolumn{2}{c|}{\textbf{CIFAR-100}} & \multicolumn{2}{c|}{\textbf{STL-10}} \\
        \hline
        \textbf{Labels/class} & 4 & 25 & 4 & 25 & 4 & 25 \\
        \hline
        (1) base   & 0.7734 & \textbf{0.9450} & 0.4117 & \textbf{0.6176} & 0.6797 & 0.8712 \\
        (2) w.mmd  & \textbf{0.9059} & 0.9371 & \textbf{0.4595} & \textbf{0.6176} & \textbf{0.7132} & \textbf{0.8782} \\
        \hline
    \end{tabular}
\end{table*}

Next, we provide an explanation of the loss function.

\[
L_{SSC}(\mathbf{Z},\mathbf{y},\boldsymbol{\lambda}) =\quad\quad\quad\quad \quad \quad\quad \quad \quad \quad \quad \quad \quad\quad \quad \quad \quad 
\]
\begin{equation}
\scalebox{1}{$
\frac{1}{\sum_{k \in \mathcal{I}} \lambda_k}
\sum_{i \in \mathcal{I}}
\frac{-\lambda_i}{|P(I)|}
\sum_{p \in P(i)}
\log \left(
\frac{\exp(\mathbf{z}^i \cdot \mathbf{z}_p) / T}{\sum_{j \in \mathcal{I} \setminus {i}} \exp(\mathbf{z}^i \cdot \mathbf{z}^j) / T)}\right)
$}
\end{equation}

Where $\mathbf{Z}$ represents the concatenation of all feature embeddings used in contrastive learning. $\mathbf{y}$ represents the labels assigned to all samples in the mini-batch, including true and pseudo labels.$\mathcal{I} = \begin{bmatrix}1, \dots, B\end{bmatrix}$ denote the set of anchor indices. The indices of samples in a mini-batch. For each anchor $i \in \mathcal{I}$, we define $P(i)$ as the set of indices of positive samples. Samples in the mini-batch that belong to the same class as $i$ but are not equal to $i$. $T$ is a temperature.
The weight $\boldsymbol{\lambda}$ represents the importance assigned to different data types. The weight allocation is determined on the basis of the type of data, such as labeled and unlabeled samples. The weight assignment strategy is described below.

\begin{equation}
\begin{alignedat}{2}
&\lambda_x = 1 \:\text{(labeled)},&\quad &\lambda_{u\uparrow}= 1\:\text{(pseudo labeled)} \\
&\lambda_c = 1\: \text{(prototypes)}, &\quad
&\lambda_{u\downarrow}= 0.2\:\text{(unlabeled)}
\end{alignedat}
\end{equation}

This weighting scheme enables the utilization of both labeled and unlabeled data for training.

\subsection{Distribution Matching}\label{match}

In this section, we describe the distribution matching between labeled and unlabeled data. First, the distance between the distributions of labeled and unlabeled data is defined as follows:
\begin{equation}
\mathbf{d}(Q_v,Q_u) = \frac{1}{m} \sum_{i=1}^m \kappa(\mathbf{v}^i) - \frac{1}{n} \sum_{j=1}^n \kappa(\mathbf{u}^j)
\end{equation}
In this context, $v$ denotes the feature vectors of labeled data, while $u$ indicates the feature vectors of unlabeled data. $Q_v$ and $Q_u$ denote the feature representations of the image data obtained after passing through the final convolutional layer. Furthermore, $\kappa$ represents a kernel function, and in our experiments, we employ the Gaussian kernel.
To encourage alignment between the two distributions, we define the MMD regularization loss as the squared norm of the distance:
\begin{equation}
    L_{mmd} =\text{MMD}(Q_v,Q_u)= \|\mathbf{d}\|^2.
\end{equation}

\subsection{Sample selection for MMD}

In the early stages of learning, feature representations are inaccurate and may not accurately capture the true distribution. Consequently, self-regularization based on early-stage feature representations may lead to the propagation of erroneous biases throughout learning.

A labeled sample $\mathbf{x}_l^i$ (and an unlabeled sample $\mathbf{x}_u^i$) is inputted into the feature extractor, producing the feature representation $\mathbf{z}_l^i$ (and $\mathbf{z}_u^i$).
The assignment probabilities $p_l^i$ and $p_u^i$ for each class are determined using the following cosine similarity equation and softmax.

\begin{equation}
\begin{aligned}
&\scalebox{1}{$p_l^i(\mathbf{z}_l^i) :=\text{softmax}({\mathbf{Z}_c \mathbf{z}_l^i})
$}\\
&\scalebox{1}{$p_u^i(\mathbf{z}_u^i) :=\text{softmax}({\mathbf{Z}_c \mathbf{z}_u^i})
$}
\end{aligned}
\end{equation}
Let $F_l$ denote the selected labeled representations and $F_u$ denote the selected unlabeled representations.
\begin{equation}
\begin{aligned}
&\text{For}\: \mathbf{x}_l^i \in \mathcal{L}\ \text{and}\:\mathbf{H}(p_l^i) \leq \epsilon_p,\:\:\mathcal{F}_l = \{ F_{\theta}(\mathbf{x}_l^i) \}
\\
&\text{For}\:\mathbf{x}_u^i \in \mathcal{U}\ \text{and}\:\mathbf{H}(p_u^i) \leq \epsilon_p,\:\:\mathcal{F}_u = \{ F_{\theta}(\mathbf{x}_u^i) \}
\end{aligned}
\end{equation}
Let $\mathcal{L} = { \mathbf{x}_l^i }$ denote the set of labeled samples and $\mathcal{U} = { \mathbf{x}_u^i}$ denote the set of unlabeled samples.
$F_\theta$ represents the feature extractor, and $\mathbf{H}$ denotes the entropy of the prediction. $\epsilon_p$ is the threshold value. Finally, the minimization of MMD is performed using the selected $F_l$ and $F_u$.

\subsection{The Loss Function}
Finally, We formulate the overall loss function by integrating the loss functions presented in \ref{semi} and \ref{match}. The resulting loss function is defined as follows:

\begin{equation}
\scalebox{1}{$
L_{total} = L_{ssc}(\mathbf{Z},\mathbf{y},\mathbf{\lambda}) +\lambda_{mmd} L_{mmd}
$}
\end{equation}
Where $\lambda_{mmd}$ controls the strength of the MMD regularization term.

\section{Experiment}

\subsection{DataSet}

In the experiments, we use three datasets: CIFAR-10 \cite{cifar}, CIFAR-100 \cite{cifar}, and STL-10 \cite{stl}. The unlabeled data in STL-10 contains unknown classes that do not belong to the ten predefined categories (airplane, bird, car, cat, deer, dog, horse, monkey, ship, and truck).

\subsection{Baseline}
As a baseline method, we employ the semi-supervised contrastive learning approach proposed in \cite{fix_con}. To demonstrate the effectiveness of distribution matching between labeled and unlabeled data, we use identical settings for all parameters except that related to distribution matching in both the proposed method and the baseline.

\subsection{Training Strategy}

For all datasets, we employed stochastic gradient descent (SGD) with momentum as the optimization method. The number of training epochs was 256, with a momentum rate of 0.9 and an initial learning rate of 0.03. The batch size for labeled data was set at 64, while that for unlabeled data was set to 448. For the learning rate schedule, we adopted cosine learning rate decay. The method for determining the learning rate $\eta_t$ is described below.

\begin{equation}
\eta_t = \eta_0 \cos(\frac{7\pi t}{16T})
\end{equation}
$\eta_0$ is the initial learning rate, $T$ is the total number of epochs, and $t$ is the current epoch.

For CIFAR-10 and CIFAR-100, we employ the Wide ResNet-28-2 architecture. For STL-10, we use Wide ResNet-37-2, as the images in the STL-10 dataset have a higher resolution compared to those in the CIFAR series. Wide ResNet-28-2 consists of 28 layers and expands the number of channels in each layer to twice that of conventional ResNet. Wide ResNet-37-2 is a deeper variant of Wide ResNet-28-2.

\subsection{Result}

The experimental results for each dataset are presented in Table 1. Except for the case of CIFAR-10, where the number of labeled data per class is 25, the method incorporating distribution matching achieves higher accuracy in all other settings.

\section{Conclusion}

From the experimental results, it was found that applying distribution matching between labeled and unlabeled data in semi-supervised contrastive learning generally improves accuracy in most cases. Furthermore, the fewer labeled samples available, the greater the benefit of distribution matching. This suggests that the proposed method is particularly useful in scenarios where labeled data is extremely scarce. On the other hand, the results on CIFAR-10 and CIFAR-100 with 25 labeled samples per class suggest that when a sufficient amount of labeled data is available, distribution matching may not lead to further improvements in learning. In future work, we plan to conduct further experiments with different amounts of labeled data to better understand the generalizability of our method.

\section*{Acknowledgement}
This work was supported in part by MEXT KAKENHI JSPS23K11174.

\end{document}